# Recyclable Waste Identification Using CNN Image Recognition and Gaussian Clustering

Yuheng Wang[*], Wen Jie Zhao[*], Jiahui Xu[*], Raymond Hong[*]

*Abstract*- Waste recycling is an important way of saving energy and materials in the production process. In general cases recyclable objects are mixed with unrecyclable objects, which raises a need for identification and classification. This paper proposes a convolutional neural network (CNN) model to complete both tasks. The model uses transfer learning from a pretrained Resnet-50 CNN to complete feature extraction. A subsequent fully connected layer for classification was trained on the augmented TrashNet dataset [1]. In the application, sliding-window is used for image segmentation in the pre-classification stage. In the post-classification stage, the labelled sample points are integrated with Gaussian Clustering to locate the object. The resulting model has achieved an overall detection rate of 48.4% in simulation and final classification accuracy of 92.4%.

*Index Terms*- Image classification, waste recycling, convolutional neural network, Gaussian clustering

## I. INTRODUCTION

In 2018, the Intergovernmental Panel on climate change detailed that global warming of 1.5 ℃ above pre-industrial levels is the limit before many natural systems to irreversible collapse [2]. Exceeding that by just 0.5 ℃ will leads to as much as 99 % declination of coral reefs that support marine environments around the world [3, 4], and a 1.5 million tons declination in global fishery catches [5]. The report details further reduction potentials for the next decades, with global reductions of fossil fuel consumption of about 80% being suggested [6, 7]. One way to aid this process is through recycling from waste, thus saving energy used in the industrial production process. Globally, waste generation has ten-folded over the past century [8]. However, only 15% of waste is recycled [9, 10, 11]. Minelgaitė et al., 2019 [12] noted that, 51.4% of the respondents who declared that they did not sort waste, demanded more convenient waste recycling facilities in the area.

Centralized waste disposal facilities [13], which collect and sort waste generated within a local or regional neighborhood area, would be a practical solution to the problem encountered in waste recycling. While the per capita municipal solid waste generation rate could be as large as 510 kg/year in European cities [14], the most crucial stage in the operation of such facilities would be the automatic classification of waste objects.

[*] University of Toronto, Toronto
27 King's College Circle, ON, Canada M5S 3G4
(E-Mail: danielwang.wang@mail.utoronto.ca;
wjz.zhao@mail.utoronto.ca; xjames.xu@mail.utoronto.ca;
raymond.hong@mail.utoronto.ca)

This study uses transfer learning from a pre-trained Resnet-50 model to generate a model which is capable of classifying images of individual waste objects into the following six categories: cardboard, glass, metal, paper, plastic, and trash. To integrate the model into actual application, which often deals with bird's-eye view of piles of waste, a sliding-window process in the pre-classification stage split the image into smaller fragments for the CNN to process, and the labelled points are integrated with Gaussian Mixture Model in the post-classification stage to locate the object.

## II. RELATED WORKS

Over the years, many works had been aimed at adding efficiency to the automatic classification of waste objects with image recognition techniques. Garcia et al., 2015 [15] had proposed a k-NN classifier which processes the RGB, grayscale and binary image of the incoming waste on a multimedia embedded processor. This classifier had achieved an efficiency high as 98.33% using parameter k = 3, however its major drawback was the inability to process multiple waste objects at once.

One of the earliest works based on machine learning was by Yang and Thung, 2016 [16]. This work compared the performance of support vector machine (SVM) and CNN in single waste object classification, and had achieved test accuracy rates of 63% (SVM) and 22% (CNN) respectively. Though the outcome seemed unpromising, this work had constructed an open source dataset, TrashNet, which is widely used in later researches.

Rabano et al., 2018 [17] proposed a transfer learning model based on Google MobileNet [18]. Using the same TrashNet in training, this work had achieved 87.2% test accuracy. Another work Costa et al., 2018 [19] used similar transfer learning technique with VGG-16 [20] and AlexNet [21], and further improved accuracy on TrashNet to 93% and 91% respectively. Going even further in this direction, Bircanoğlu et al., 2018 [22] provided a detailed comparison of the performance of multiple well-known CNN architectures for transfer learning on TrashNet, including ResNet50 [23], Inception-v4 [24], and DenseNet [25]. This work had achieved 95% test accuracy on a fine-tuned DenseNet model.

Other than single object classification, a different approach was discussed in Awe et al., 2017 [26], which proposed to use faster region-based CNN (Faster R-CNN) to categorize mixed waste object piles into three categories (i.e. landfill, recycling and paper) without singling out individual objects. Another more simplified approach was done by Mittal et al., 2016 [27], developing a smartphone application only to detect garbage objects in images using transfer learning from AlexNet.

## III. AUGMENTED TRASHNET DATASET

The open-source TrashNet dataset is augmented for the training of the model. The original dataset contains 2527 RGB waste images of size 512*384 pixels. Each image is taken under these conditions: object being located at the center, with white background, common indoor lighting condition and no occlusion shadow. The images are labeled with 6 categories (paper, glass, trash, metal, plastic, cardboard) Fig. 1 shows sample waste images from the original dataset representing each class:

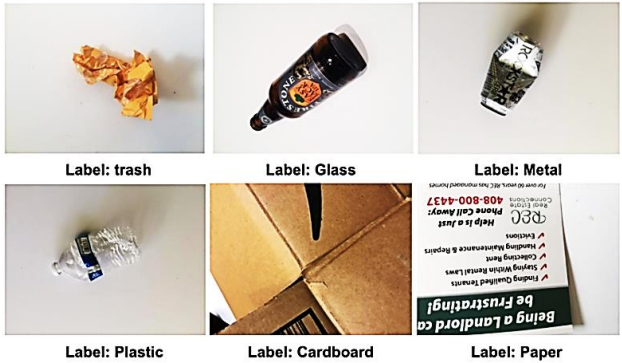

Fig. 1. Sample images from original TrashNet dataset

The original dataset is not evenly distributed. There are 594 images labelled *Paper*, 501 images labelled *Glass*, 482 images labelled *Plastic*, 410 images labelled *Metal*, 403 images labelled *Cardboard* and 137 images labelled *trash*. Since an evenly distributed dataset is desirable in training, a portion of original images were randomly cropped and flipped (See Fig. 2) to augment the dataset to 3600 images of size 170*128 pixels, with 600 images of each label.

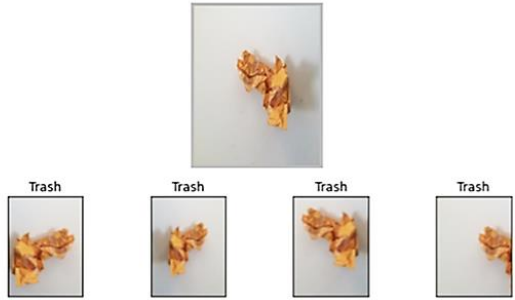

Fig. 2. Example of original images and processed images

## IV. METHODOLOGY

### A. Retraining Resnet-50 CNN model for classification

The transfer learning model consists of a pretrained Resnet-50 convolutional neural network and a fully connected layer. The Resnet-50 CNN is pre-trained with ImageNet dataset [28]. As shown in Fig. 3, the Resnet-50 CNN was used to do feature extraction, while the fully connected layer acts as a classifier to make the final classifications. Only the fully connected layer is trained with augmented TrashNet dataset.

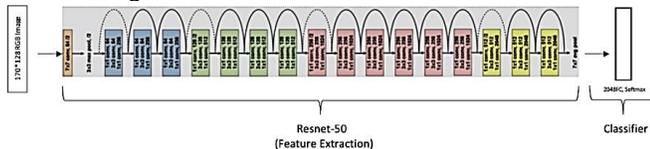

Fig. 3. Transfer learning model architecture

The model was trained with images randomly selected from the training set for 140 iterations. Multiple sets of hyper-parameters were traversed and the optimal setting was chosen, with the learning rate of 0.0001 and the epoch number of 30. The model has achieved a final training accuracy of 98.8%, validation accuracy of 88.9% and test accuracy of 92.4%. Fig. 4 shows the accuracy over iterations curves on training and validation set separately, and Fig. 5 shows the training loss over iterations curve.

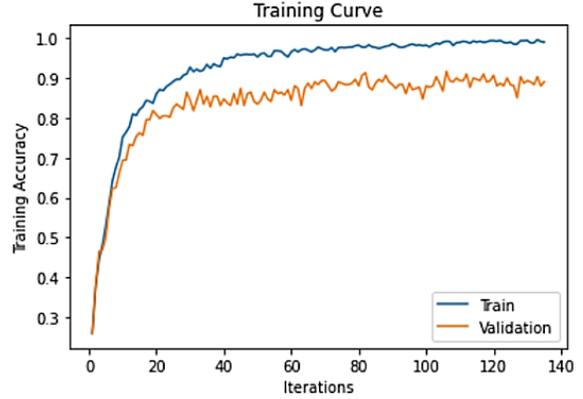

Fig. 4. Training and validation accuracy curve

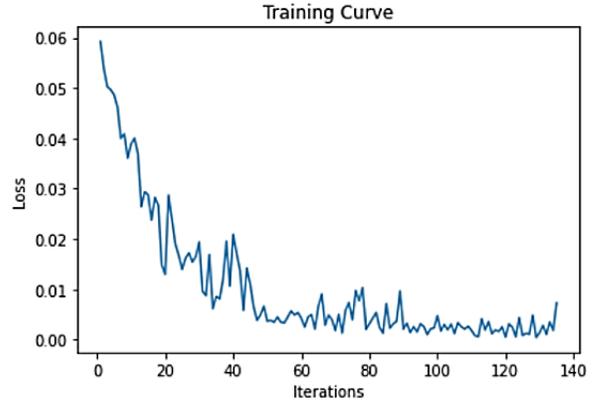

Fig. 5. Training loss curve

### B. Observation from classifier model

Fig. 6 is the confusion matrix of the testing set, which gives insight of the model's relatively poor performance in distinguishing among *Metal*, *Plastic* and *Glass* objects.

|  | Cardboard | Glass | Metal | Paper | Plastic | Trash |
|---|---|---|---|---|---|---|
| Cardboard | 54.10% | 5.88% | 1.17% | 16.47% | 11.76% | 10.58% |
| Glass | 0% | 65.32% | 6.45% | 8.06% | 10.48% | 9.67% |
| Metal | 2.43% | 21.95% | 48.78% | 10.97% | 9.75% | 6.09% |
| Paper | 0% | 3.67% | 2.20% | 87.49% | 5.14% | 1.47% |
| Plastic | 1.01% | 20.20% | 7.07% | 10.10% | 56.56% | 5.05% |
| Trash | 0% | 1.12% | 0% | 0% | 0% | 98.87% |

Fig. 6. Confusion matrix of the testing set

A conjecture of the reason is that the three categories are all highly reflective materials, and appear to be similar to each other under strong lighting conditions. A further attempt to manually reduce the contrast ratio and/or the brightness has significantly reduced the false-positive rate of *Glass* and *Metal*, and increased the true-positive rate of *Plastic*, while having minor effect on the true-positive rate of *Glass* and *Metal*, thus improve the performance.

*C. Sliding-window*

The Resnet-50 CNN classifier model requires the input to be the image of a single waste object. However, in the real-world application, due to the placement of camera and its distance to the waste pile, the images are usually in bird's eye view of multiple waste objects. Therefore, a preliminary stage of splitting the image to extract single waste objects is necessary.

Fig. 7 demonstrates a sliding-window process by traversing a large-scale image with fixed-size windows, and generating multiple image segments in sequence. The tunable parameters include the window's size and the step-length, i.e. the overlapping rate of two consecutive windows.

When each image segment is generated, its position in the original image's reference frame is also recorded to form an image-coordinate pair as the input of the classifier model.

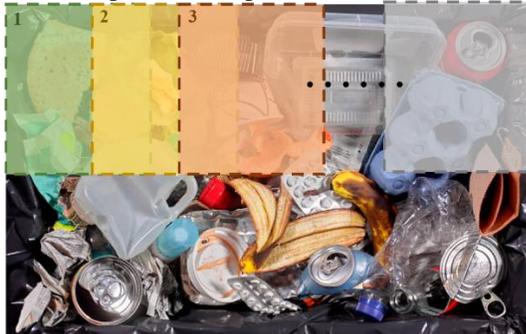

Fig. 7. Sliding-window process

*D. Gaussian clustering*

After the classifier model has labeled an image-coordinate pair, a corresponding label-coordinate pair is generated. The final object detection is produced by performing Gaussian clustering on those label-coordinate pairs. The position and size of the waste object is determined based on the mean and variance of the Gaussian cluster respectively. The number of Gaussian clusters is determined based on the scale of the original image, which should be tuned to match the average number of waste objects in a typical waste pile (See Fig. 8).

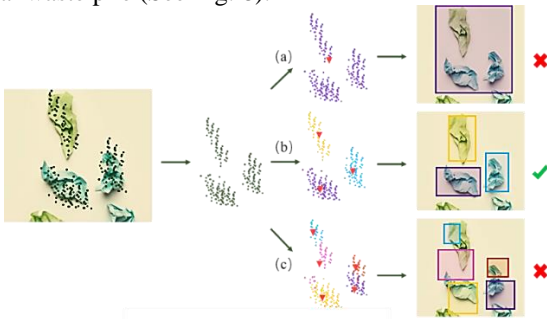

Fig. 8. Effect of different number of Gaussian clusters: (a) too few, (b) just fit and (c) too much

## V. EXPERIMENTAL RESULTS

Fig. 9 demonstrates a simulation of solving actual waste classification problem with the proposed model. Table 1 shows the quantitative analysis based on the result of the simulation. The false-positive rate of *Glass* objects is high, while the false-negative rate of *Plastic* and *Metal* objects is low. It was hypothesized that the strong lighting condition in the image had interfered the classification of *Glass* and *Plastic* objects, and most of the *Metal* objects that had escaped the detection are too small in size. The overall detection rate is 48.4%.

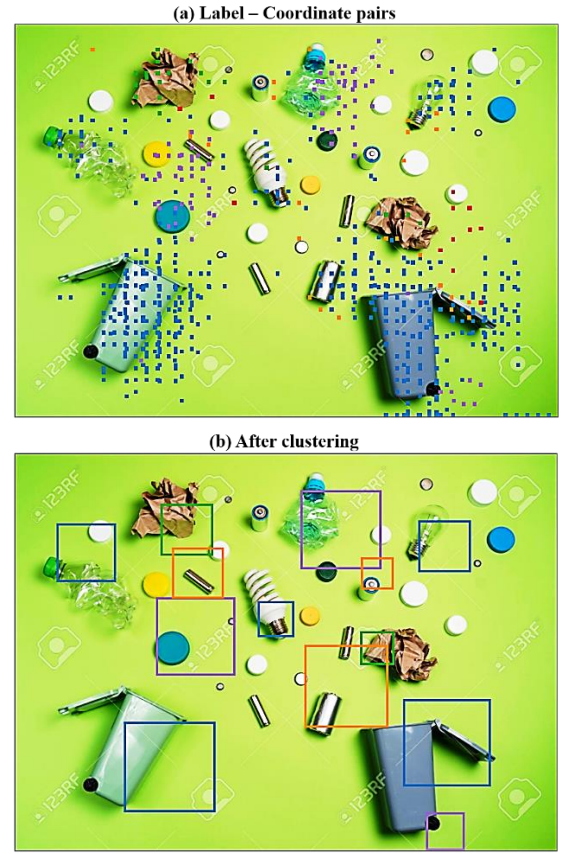

Fig. 9. Simulation of a waste detection problem. In visualization, different color represents different labels: red for *Cardboard*, blue for *Glass*, orange for *Metal*, green for *Paper* and purple for *Plastic*, *trash* category is filtered out in visualization

| Label | Correctly Identified | Identified | Total |
|---|---|---|---|
| Cardboard | 0 | 0 | 0 |
| Glass | 2 | 5 | 2 |
| Metal | 3 | 3 | 11 |
| Paper | 2 | 2 | 2 |
| Plastic | 5 | 5 | 16 |

Table 1. Quantitative analysis of the simulation result

## VI. DISCUSSION AND OUTLOOK

Most of the previous works had been focusing on the identification of single objects. While this study has maintained high performance in identification, it also combines the effort of mass detection to process images of mixed waste objects. However, it is necessary to rethink the setting of the model before deployment in centralized waste disposal facilities. The most salient problem is the dependency on optical conditions.

Even in the case of human kind, one would often need other sensations like auditory or tactile. Tarbell et al., 2010 [29] details a model that distinguishes between plastic, metals and glass objects through attributes not only optical but also acoustic characteristics. While this solution exceeds the scope of this model, future studies could be conducted on this topic.

Further optimization of the Gaussian clustering could follow the pathway described by Neagoe and Chirila-Berbentea, 2016 [30], in which a method of combining the result of K-cluster and Gaussian mixture model, and using expectation-maximization technique to enhance the robustness, has been proposed and thoroughly studied.

Waste classification turns out to be a complex problem with extensive relevant factors. Unlike other problems dealing with simplified images or numeric data, this problem needs to consider the object in relation to its surroundings. The model would have to determine the number of objects, to identify the boundary between distinct objects, and to classify images under suboptimal lighting conditions, overlapping positions, and uneven sizes. The model also has to avoid the most common mistakes in image recognition, including misidentifying multiple distinct objects as one single object or vice versa, repeatedly identifying the same object in different classes, and failure to identify hidden objects.

## VII. Conclusion

In this study, a classifier model based on transfer learning from Resnet-50 CNN is successfully developed using PyTorch. After training with the augmented TrashNet dataset, the model has achieved classification accuracy of 92.4% and a detection rate of 48.4% on a large-scale mixed waste object image.

## Acknowledgment

This work is supported by the University of Toronto's Edward S. Rogers Sr. Department of Electrical and Computer Engineering. Computational resources were provided by the university's Engineering Computing Facility.

## References

[1] garythung, "garythung/trashnet," GitHub. [Online]. Available: https://github.com/garythung/trashnet. [Accessed: 31- Aug-2020].

[2] "Global Warming of 1.5 ºC —", Ipcc.ch, 2020. [Online]. Available: https://www.ipcc.ch/sr15/. [Accessed: 31- Aug2020].

[3] K. Frieler, M. Meinshausen, A. Golly, M. Mengel, K. Lebek, S. Donner, and O. Hoegh-Guldberg, "Limiting global warming to 2ºC is unlikely to save most coral reefs," Nature Climate Change, vol. 3, pp. 165–170, 02 2013.

[4] C.-F. Schleussner, T. Lissner, E. Fischer, J. Wohland, M. Perrette, A. Golly, J. Rogelj, K. Childers, J. Schewe, K. Frieler, M. Mengel, B. Hare, and M. Schaeffer, "Differential climate impacts for policy-relevant limits to global warming: The case of 1.5ºC and 2ºC," Earth System Dynamics Discussions, vol. 6, pp. 2447–2505, 11 2015.

[5] W. Cheung, G. Reygondeau, and T. Frolicher, "Large bene-¨ fits to marine fisheries of meeting the 1.5°c global warming target," Science, vol. 354, pp. 1591–1594, 12 2016.

[6] Z. Klimont, K. Kupiainen, C. Heyes, P. Purohit, J. Cofala, P. Rafaj, J. Borken-Kleefeld, and W. Schoepp, "Global anthropogenic emissions of particulate matter including black carbon," Atmospheric Chemistry and Physics, vol. 17, pp. 8681–8723, 07 2017.

[7] A. Stohl, B. Aamaas, M. Amann, L. Baker, N. Bellouin, T. Berntsen, O. Boucher, R. Cherian, W. Collins, N. Daskalakis, M. Dusinska, S. Eckhardt, J. Fuglestvedt, M. Harju, C. Heyes, Hodnebrog, J. Hao, u. im, M. Kanakidou, and T. Zhu, "Evaluating the climate and air quality impacts of short-lived pollutants," Atmospheric Chemistry and Physics, vol. 15, 09 2015.

[8] I. Zelenika, T. Moreau, and J. Zhao, "Toward zero waste events: Reducing contamination in waste streams with volunteer assistance," Waste Management, vol. 76, 03 2018.

[9] A. Zaman, "A comprehensive study of the environmental and economic benefits of resource recovery from global waste management systems," Journal of Cleaner Production, vol. 124, 02 2016.

[10] N. Pietzsch, J. L. Ribeiro, and J. Fleith de Medeiros, "Benefits, challenges and critical factors of success for zero waste: A systematic literature review," Waste Management, vol. 67, 05 2017.

[11] H. Jouhara, D. Czajczynska, H. Ghazal, R. Krzy´zy´nska, ´L. Anguilano, A. Reynolds, and N. Spencer, "Municipal waste management systems for domestic use," Energy, vol. 139, 07 2017.

[12] A. Minelgaite and G. Liobikiene, "The problem of not waste ˙ sorting behaviour, comparison of waste sorters and nonsorters in european union: Cross-cultural analysis," Science of The Total Environment, vol. 672, 03 2019.

[13] T. Eichner and R. Pethig, "Recycling, producer responsibility and centralized waste management," FinanzArchiv Public Finance Analysis, vol. 57, pp. 333 – 360, 02 2000.

[14] P. Beigl, G. Wassermann, F. Schneider, and S. Salhofer, "Forecasting municipal solid waste generation in major european cities," 06 2004.

[15] A. Torres Garcia, O. Rodea-Aragon, O. Longoria-Gandara, ´F. Sanchez-Garc´ıa, and L. Gonzalez-Jim´enez, "Intelligent ´ waste separator," Computacion y Sistemas, vol. 19, pp. 487– 500, 09 2015.

[16] Yang, Mindy and Thung, Gary, "Classification of trash for recyclability status," arXiv preprint, 2016.

[17] S. Rabano, M. Cabatuan, E. Sybingco, E. Dadios, and E. Calilung, "Common garbage classification using mobilenet," pp. 1–4, 11 2018.

[18] Google, "MobileNets: Open-Source Models for Efficient On-Device Vision," Research Blog. [Online]. Available: https://research.googleblog.com/2017/06/mobilenetsopen-sourcemodels-for.html. [Accessed: 08-Dec-2017].

[19] A. Silva and E. Soares, "Artificial intelligence in automated sorting in trash recycling," 10 2018.

[20] K. Simonyan and A. Zisserman, "Very deep convolutional networks for large-scale image recognition," arXiv 1409.1556, 09 2014.

[21] A. Krizhevsky, I. Sutskever, and G. Hinton, "Imagenet classification with deep convolutional neural networks," Neural Information Processing Systems, vol. 25, 01 2012.

[22] C. Bircanoglu, M. S. Atay, F. Beser, O. Genc, and M. A. ˘ Kizrak, "Recyclenet: Intelligent waste sorting using deep neural networks," 07 2018.

[23] K. He, X. Zhang, S. Ren, and J. Sun, "Deep residual learning for image recognition," pp. 770–778, 06 2016.

[24] C. Szegedy, S. Ioffe, V. Vanhoucke, and A. Alemi, "Inception-v4, inception-resnet and the impact of residual connections on learning," AAAI Conference on Artificial Intelligence, 02 2016.

[25] G. Huang, Z. Liu, L. van der Maaten, and K. Weinberger, "Densely connected convolutional networks," 07 2017.

[26] Awe, Oluwasanya and Mengistu, Robel and Sreedhar, Vikram, "Smart trash net: Waste localization and classification," arXiv preprint, 2017.

[27] G. Mittal, K. Yagnik, M. Garg, and N. Krishnan, "Spotgarbage: smartphone app to detect garbage using deep learning," pp. 940–945, 09 2016.

[28] ImageNet. [Online]. Available: http://www.image-net.org/. [Accessed: 01-Sep-2020].

[29] K. A. Tarbell, D. K. Tcheng, M. R. Lewis, and T. A. Newell, "Applying machine learning to the sorting of recyclable containers," 2010.

[30] V.-E. Neagoe and V. Chirila-Berbentea, "Improved gaussian mixture model with expectation-maximization for clustering of remote sensing imagery," pp. 3063–3065, 07 2016.